# Position and Vector Detection of Blind Spot motion with the Horn-Schunck Optical Flow

**Stephen S. Yu**
Torrey Pines High School
San Diego, CA 92130
istephenyu@gmail.com

**Mike H. Wu**
Torrey Pines High School
San Diego, CA 92130
grub007@gmail.com

*\*Acknowledgements extends to Dr. Brendan Morris of the University of Nevada, Las Vegas;
Professor Paul Yu of the University of California, San Diego*

## ABSTRACT

The goal of the research is to create a cost-effective way to detect the presence of a vehicle in the blind spot in real time by computing its position and relative motion. To do, we adapted the Horn-Schunck Optical Flow method [1], which traces pixels onto the next frame and draws vector arrows representing the movement of the pixels[4]. The vectors were hen categorized into car and background vectors.

For position detection, an *box*-algorithm was developed to localize the position of the car. The center of the box is the average of the (x,y) coordinate of the car vectors while the size of the box is directly proportional to the ratio of car to background vectors. An additional feature was the use of stereo-cameras that extract data from duo cameras positioned at converging angles to visualize depth of motion in specific cases of multiple object differentiation in the blind spot. As a confirmation technique, circular object detection was used to verify the accuracy of optical flow analysis. An edge detection was used before the circle detection to draw the layouts of the vehicle. A distance threshold was applied to the stereo data to remove background objects.

In the trials that ran, the program was able to accurately detect the vehicle and track its motion throughout the video. The code developed also extended to bicyclists, pedestrians and motorcyclists. Furthermore, the code was not depended on sunlight. Thus, if applied to the real world, the real-time capabilities could be revolutionary in promoting safer driving conditions.

## Categories and Subject Descriptors
- Image Processing
- Robotic Intelligence Systems

## Keywords
- Optical Flow
- Edge Detection
- Hough Transform
- Box Capture

## Introduction

Blind spot negligence often leads to unsafe lane changes and dangerous turns which, in turn, can cause crashes and injuries. In 2008, the Seattle Auto Pacific Center published statistics speculating that about 400,000 of the car accidents occurring each year were due to blind spot negligence (Figure 1). The two most prominent solutions provided are convex shaped mirrors and high-tech radar sensors[5]. Though the mirror minimizes blind spots, because it does not give a full, 360 degree view, it cannot completely eliminate blind spots and therefore does not solve the issue. The radar method works efficiently and has been implemented with some success; however because this radar technology[5] is expensive it is generally only employed in luxury vehicles, and may not be appealing, or even affordable, to the average citizen. Thus, the research goal had to be cost-effective; in this case the goal was achieved with optical cameras. The hypothesis was based on an algorithm called optical flow (Figure 2, 3). Because video cameras are capable of producing individual frame images, and pixels within the images move between frames, generation of flow vectors to track motion seemed plausible. If that was possible, then the vehicle vectors could be isolated, and the presence of any object could be detected. Optimally, a product could be constructed so that objects entering a vehicle's blind spot would cause a light placed on the wing mirror (Figure 4) to flash a corresponding color, alerting the driver not to make, for example, a lane change or a sudden turn. The product should help drivers make the right decisions when operating a vehicle and help prevent a fraction of accidents because incoming objects can be detected early. The system also has possible applications in other fields of study. For example, consider unidentified aerial vehicles (UAVs). A tracking module using a video camera can give aircrafts the ability to follow or find UAV's based visual stimuli.

## Materials and Methods

The general experiment was split into two areas: one completed for real-time video capturing and the other for video processing through a computational program. For the video capturing, the scope of the project required two types of videos, a regular 2-D frame and a black and white stereo video (Figure 18). The regular input was obtained by pointing a camera recorder towards the vehicle's blind spot and recording any objects that passed by at different angles. However, the stereo video required more sophisticated technology and was done with dual lens equipment (stereo video refers to image frames whose pixel intensity represents distance from focal lens (Figure 18)). From there, the video was run in our written Matlab program that processed the video to find motion in specific scenarios based on a series of algorithms. An adapted version of optical flow by Horn and Schunck[1] was applied to the system in all cases, regardless of the situation. Optical flow[1] was then edited to view two subsequent frames of the video and pixel movement between those frames was compared[4]. More specifically, each pixel was set with a relatively unique brightness factor compared to the pixels around it, and optical flow could track that brightness onto the next frame to find the same pixel[4]. The length moved was calculated and a vector arrow was drawn to represent that motion. From there, the vectors were split into two colors: green for forward object (car) motion and red for opposite background motion. The car vectors were later assumed to hold an angle of $\pi/4$ (Figure 6); everything outside of this threshold was defined as a background vector, eliminating many miscalculated green vectors (Figure 6). Since the object was approaching from the rear, the assumption could be made that the object vectors would be heading to-the-right. Later, specific adjustments were made to limit the error range. Once the flow image was generated, different algorithm paths were taken based on circumstance.

For the common scenario where an object approached the vehicle from the rear blind spots(s), the goal was to generate a blue box to essentially follow the object (Figure 7). However, first, the presence of an object had to be confirmed. The ratio of object vectors to background vectors, based on a circular histogram (rose plot) (Figure 13), was looked at. If the ratio surpassed a threshold of 0.1, then further calculations were made. Otherwise, the frame was skipped to conserve calculation time. If the first threshold test passed, then the construction of the box began. Note that each vector arrow could be quantified with an (x,y) coordinate. Thus, the initial center of the box was the average of all the (x,y) coordinates of all the green object vectors. Building on that, the size of the box should fluctuate based on the logic that as an object (car) approached its size seemed to increase in the driver's point of view (Figure 7) (as well as the camera's). When the object got closer, it generated more green vectors and blocked out more red vectors, increasing the ratio, so the size of the box was set proportional to the ratio between the object and background vectors: having a changing box size allowed for a more accurate prediction of the object's position. The next step was to revisit the error range. The initial assumption made included a lot of object vectors that were actually part of the background. To solve this problem, a standard deviation filter was added for a more accurate box center (Figure 8): From the initial box center, any object vectors three standard deviations away were removed and the center was recalculated. The process was repeated with two standard deviations (stddevs) and then one stddev in a cycle until the final box center was as accurate as possible. At this step, a solid image was generated with a simple motion detector; from here, it was advantageous to program a series of safety mechanisms to prevent any detection of "ghost" motion and increase the system's reliability. The most prominent one used was a second threshold test. After the box was generated, if 50% of the object vectors were not located inside the box, then the box was nullified based on insufficient evidence and the frame was discarded. A number of movement thresholds were applied, mostly based on the Pythagorean Theorem to limit how far the center of the box could move within a given time frame (See figures 9, 10, 11, 12 for details on thresholds used). For example, if the center started out with coordinates in the far right of the frame, it was unlikely for that center to teleport to the far left within a couple frames. Thus, any sudden movements were discarded as noise meaning and no random small boxes would appear at the edges of the screen, minimizing false detection.

The basic scenario described above worked for most blind spot cases and would have been a sufficient motion detector except in two cases: when an object was in front of the vehicle and entering the frontal blind spot and when an object entered the blind spot and remained there at around a relative velocity of about zero.

Objects decelerating into the blind spot from the front created a problem because the deceleration vectors blended into the background and nothing was detected. Thus, a completely alternate system based on shape detection was required. Given that all vehicles, ie) bicycles, motorcycles, cars share the same shape of wheels, Hough Transform and edge detection[3] (Figure 16, 16b, 17) could be combined for circle detection. Given any image frame, edge detection[3] (Figure 17) accelerates the rate of computation many fold by removing any excess pixels and only keeping the crucial shape outlines. Given these outlines, Hough Transform (figure 16, 16b) was set to operate off of a voting system: every pixel generated a circle of radius r; the area with the most overlapping lines was voted by its surrounding pixels to be the center of a present circle (in this case, the location of the wheel). If two or more circles were detected, than the light in the wing mirror (figure 4) shines to alert that an object is present. The same idea applies with objects of no relative motion (figure 20). For example, when a car stays fixated in the blind spot, the pixels generated are not sufficient to surpass the magnitude threshold (not enough object vectors) and the frame is discarded even though a car is present. Using the circle detection method, as the car approaches, even though it disappears in terms of flow vectors, its wheels are stationary, allowing for its detection.

The next scenario to be considered is quite relevant to real life road travel. Very rarely on the road is there only one car in the blind spot at once; usually, many cars filled the multiple lanes (figure 19), especially during the heavy traffic hours. The current system is insufficient in counting the number of cars in the blind spot since it can only detect the presence of motion, not the quantity. Thus, instead of placing a single camera on the wing mirror, two linked cameras lenses were placed a specific distance apart for stereo vision (figure 18). The two camera lenses acted similar to the human eyes, in that their foci's meet at a converging point, giving a sense of distance in terms of pixel units. The result was the generation of an image of the video frame in black and white where the degree of brightness of the pixel was directly proportional to distance away from the lens (white being further away and black being closer). The image could also be rendered in color schemes where blue represented greater distance and red/orange represented closer distances. By using the correct threshold ranges, it was possible to isolate objects one lane away, two lanes away, etc so that the driver could know exactly how imminent the danger was (figure 19). Adding onto that idea, the alert system was then made into a triple layered light alarm (figure 4). When an object was in the next lane, a solid red light would shine, signifying imminent danger. When an object was two or three lanes away, a solid yellow light would shine, signaling potential, but not immediate, danger. Finally, when no object was present or when an object was greater than three lanes away, a green light was shined.

It is worthy to note that although it is ideal to use real time video data for the coding processes, certain scenarios call for nearly unrealistic situations. For example, in the case of no motion when the approaching object is traveling at exactly the same speed at the vehicle, which is highly difficult to capture on tape. As a result, a Python-based Blender 3D simulation system (figure 5, 12, 20) was used that contained a user interface that allowed for manipulation of objects and environments. Using this program, specific velocities were set for various objects (of multiple cars in the blind spot, an object entering from both directions and an object traveling at the same speed) which could be easily produced and exaggerated to test the limits of the code. (The speeds were decided based on the relative velocities of the armature/camera compared to the actual car object). An important note on this is that in order for the Blender generated videos (figure 5, 12, 20) to function properly, the background had to be manually imputed. A copy of simulated trees, sky, and grass was created and used in the simulation in order to produce proper background vectors. (A regular stationary gray screen is unsatisfactory since there is no brightness contrast between surrounding pixels.) Alternatively, before such a resource was available, inputting several randomly generated shapes in the background helped create a portion of vectors that surpassed the threshold tests.

## Results and Analysis

Because the project was the design of a product, the results were based on accuracy and efficiency: by running the code designed in numerous situations and in many (9-10) trials of <u>each</u> situation, we attempted to uncover whether the system was ready for development. Analysis of the system's performance was rather straightforward; the code was programmed to process a video and generate an edited video with the vectors/box and also print a short message whenever an object was detected. If the box generated actually encompassed an object and no random error messages were being printed in a multitude of trials, then it could be generalized that the program was working sufficiently

When dealing with normal vehicles approaching from behind, the code works very well (Figure 13). In the all of the trials run, the system was able to detect the motion of an object and alert with a simple Matlab message. In fact, the Optical Flow implementation worked a great deal better than any other optical methods (like pure shape detection per-say) because an incoming car was detected right from the beginning of its exterior even before, for example, the wheels appeared for shape detection. Another key aspect to the designed system was that it was able to detect an object regardless of its shape because it was based off of motion vectors implying that it can be applied to detecting bicyclists, motorcycles, pedestrians (Figure 15), etc. This generic accuracy is very applicable to real time driving because objects need to be detected regardless of what they are. Furthermore, the source code proved to be successful even when there was a lack of light, such as during the night (Figure 14). The headlights of the car proved to generate enough car vectors for the code to create the box and surround the object. The surrounding environment (red vectors) was generated by lamp posts or street lights. Four trials were done in night (figure 14) and all four proved to be successful, in the sense that the car(s) were detected. However because the source code is unable to function properly without light we suspect that given more night trials, sooner or later, one will fail. In low light situations, and situations in which no background vectors were present, small fluctuations in movement, even ones in which the magnitude was lower than the threshold, were detected.

In the more obscure cases in which a higher level of methodology was required, the accuracy of the results decreased: when shape detection was necessary, occasionally, circles were plotted that did not characterize motorized wheels. For example, random circles were detected in the sky or the walls of a building, circles that could cause potential accidents. Because the accuracy of the system was a bit lower than the vector algorithm, further thresholds were necessary in order to limit circle flexibility. The wheels of a car only appear in certain areas of a video frame, no matter where the road merges or splits; in other words, the extreme corners could be nullified from calculation because it was illogical for any relevant motion to appear there. By splitting the screen into regions of interest versus non-interest, faulty detections could be ignored and the accuracy of the shape detection system multiplied.

The stereo application (figure 18) was an exciting addition because it expanded the abilities of the source code. No other developed program so far is able to distinguish the distance of approaching objects and able to calculate the number of objects present with only a visual stimulus. Although the brightness of the stereo images isn't in real meters, the luminescence factor provides enough information to apply the right thresholds so that the code can print relative distance between vehicles and number of vehicles. Not only that, but also the way that the stereo algorithm processes images formats most of the background to darker blue pixels due to the distance away from the converging cameras. This allows for a separation between background and foreground because all the darker blue pixels can essentially be eliminated without fear since even if a car was present in that area, it is too far away to cause concern. Taking away a portion of each frame allowed for a clearer representation of the object of interest and allowed for an increase in processing speed since there was less to process.

Though the system has many positives, the downfalls of the system must also be considered. The most prominent issue is the amount of processing time that a video requires. In the case of a radar sensor system[5], the feedback is almost instantaneous and continuous since a constant stream of waves is being emitted. However, the optical technique proposed requires an Intel i3 processor to be in real time. Before even considering hardware, the actual Matlab program is an interpreter, not suited for mass computation; when the code was transferred to C++, the speed dramatically increased. In addition, small edits to the source code here and there cut off a good deal of the time required to process 100 frames. One such edit was made because the actual vector image and circle detection images need no frame image; the output file only needed to be pure arrows or circles. In addition it was not necessary to read every single frame, even processing every 5 frames proved to be sufficient to detect the vehicles to a respectable accuracy. In regard to hardware, instead of a serial processing unit, parallel processing with MPI or even parallel GPU processing with CUDA can be applied.

## Diagrams

Captions are included with the visual images that are referred to in the text. Note that because this project is almost entirely based on a visual diagnosis, this section may be longer than most

### General Information:

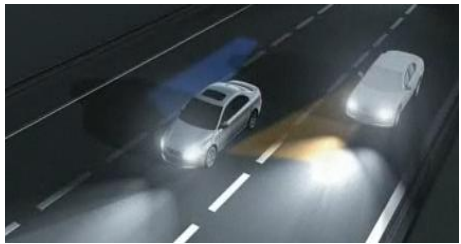

**Figure 1**: The orange and blue shaded areas Represent the blind spots of interest

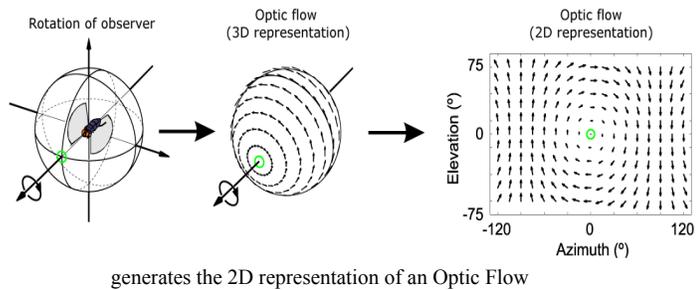

generates the 2D representation of an Optic Flow

**Figure 3**: The calculations of Optic Flow, notice that u and v are flow vectors. The I represents intensity derivatives.

$$E = \iint \left[(I_x u + I_y v + I_t)^2 + \alpha^2 (|\nabla u|^2 + |\nabla v|^2)\right] \mathrm{d}x\mathrm{d}y$$

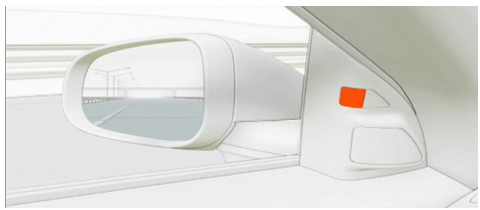

**Figure 4**: Diagram of the proposed alert system With red light = immediate danger

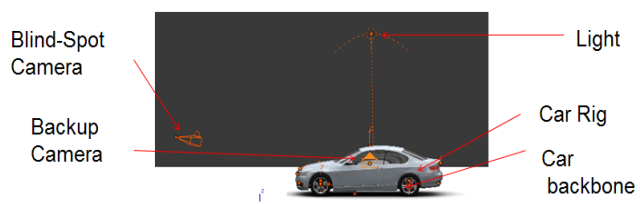

**Figure 5**: An example of a picture made in Blender 3D. The important parts are the Car rig and the camera.

### Visual Methodologies for Box capturing:

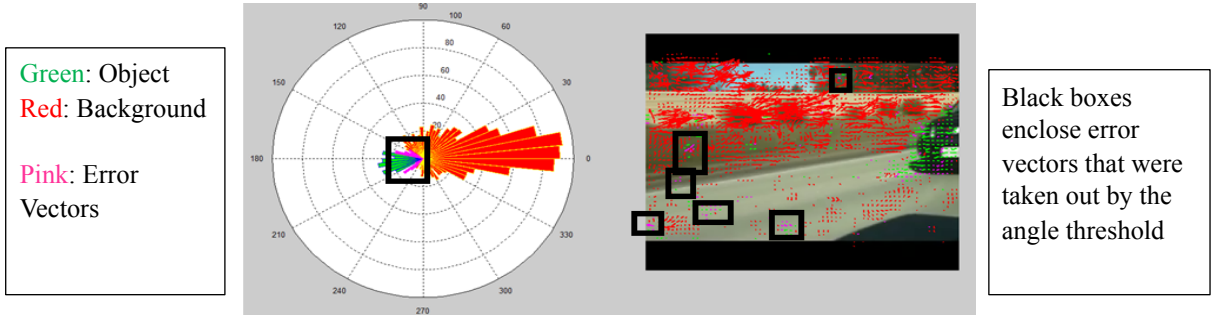

Green: Object
Red: Background

Pink: Error Vectors

Black boxes enclose error vectors that were taken out by the angle threshold

**Figure 6**: Angle Threshold: The diagram on the right is the quiver plot while the diagram on the left is the rose plot. Since we needed to distinguish the car vectors from the background vectors, we made an angle threshold filter with the car vectors having a range of π/4. The pink bars in the rose plot and the black boxes in the quiver plot show vectors that were converted to background from car vectors. Notice that this increases accuracy.

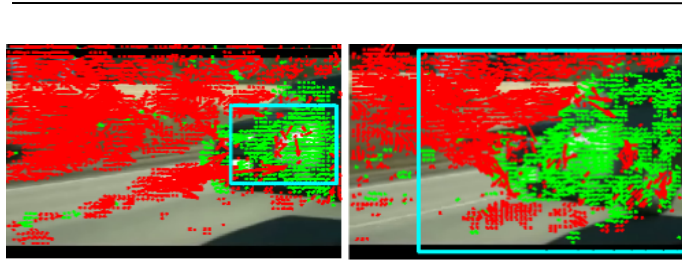

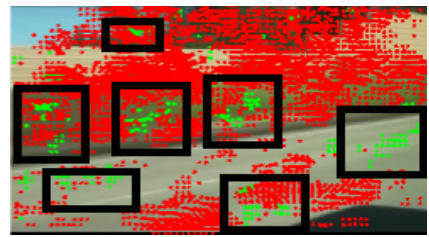

Black boxes contain filtered out vectors

**Figure 7**: Notice that the box gets bigger according the Ratio between green and red vectors, meaning bigger box for closer car.

**Figure 8**: The standard deviation filter is used to make the center of the blue box more accurate. The black boxes show computational errors that are removed with the filter.

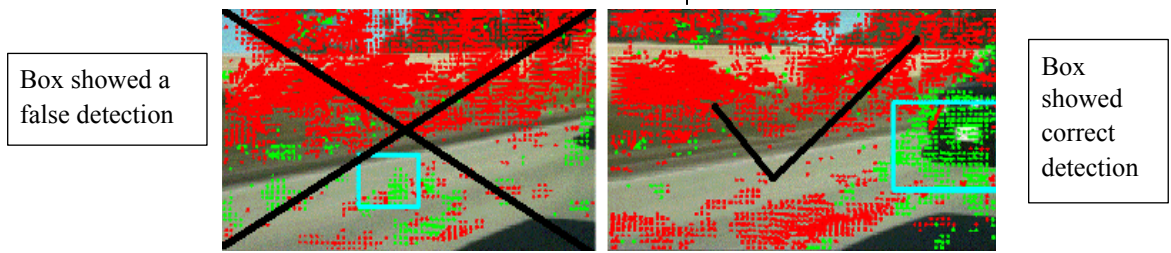

Box showed a false detection

Box showed correct detection

**Figure 9**: The movement threshold is used to prevent small boxes from forming at the edge of the video screen. The algorithm is that a box center cannot move too much within 5 frames = threshold.

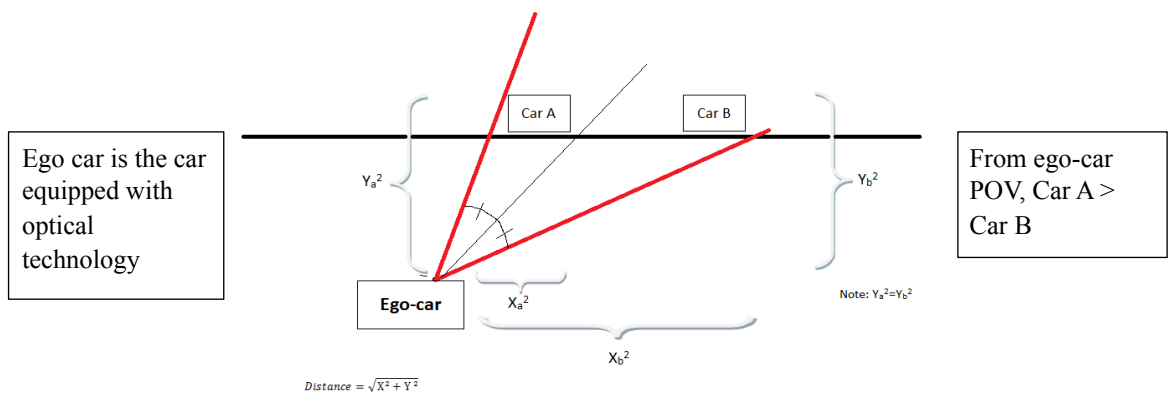

Ego car is the car equipped with optical technology

From ego-car POV, Car A > Car B

$Distance = \sqrt{X^2 + Y^2}$

**Figure 10**: Pythagorean Nullification – Given two frames, X = horizontal distance, Y = vertical distance. Since $X_a^2 < X_b^2$, and $Y_a^2 = Y_b^2$, then distanceCar A<distanceCar B, meaning that Car A will appear bigger than Car B. Thus, you can ignore small boxes on left side of frame = better accuracy.

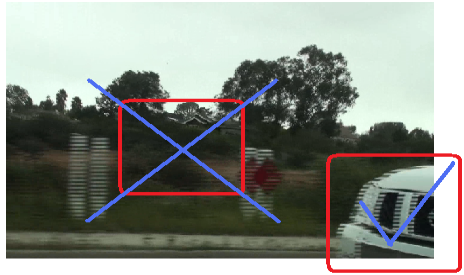

**Figure 11**: False Box Safety Nullification – If a car is approaching from the side, a box should be formed first from the left and right sides (because cars rarely pop out from nowhere), thus eliminate boxes that appear in the middle.

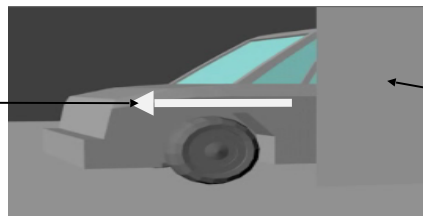

**Figure 12**: False Box Safety Nullification (part 2): There is one exception to the rule: Cars can appear out of nowhere if there's a wall covering them initially. Thus, keep a box that appears in the middle only if a large enough amount of vectors suddenly appear.

Vector Data Collection/Analysis:

Green: Object Vectors
Red: Background Vectors

Notice Box changes size to match vector ratio (rose plot)

Normal Situation:

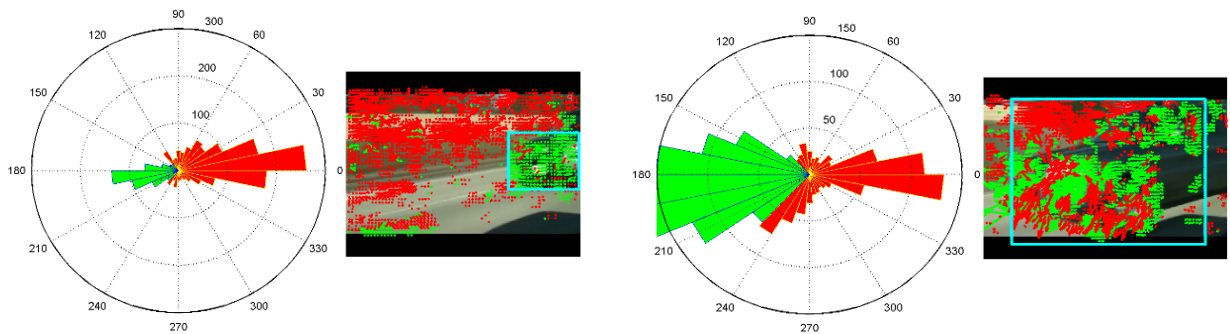

**Figure 13**: In the circular plots, the green bars represent the car vectors while the red bars represent the background vectors. The height of the bars represents the quantity of vectors not the magnitude. Notice that they are separated based on angle. In the quiver plots (the box-like ones), there are again the red and green vectors. Notice that this time there is a blue box surrounding the car. The image to the right has a bigger box because the car is closer.

Nighttime Situation:

Green: Object Vectors
Red: Background Vectors

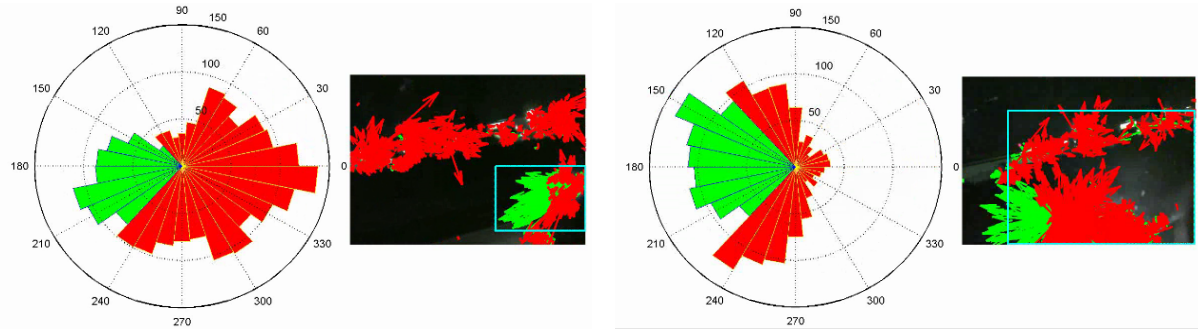

**Figure 14**: Notice that there is amassed amount of red arrows towards the upper regions of both of the quiver plots, as confirmed by the circular rose plots. Those are generated from the lamp posts and the headlights of the opposite cars. Notice that the only green arrows are from the headlights of the car of interest. Notice that bar is still able to enlarge as the car moves closer.

## Variable Objects (not cars) Situation:

Green: Object Vectors
Red: Background Vectors

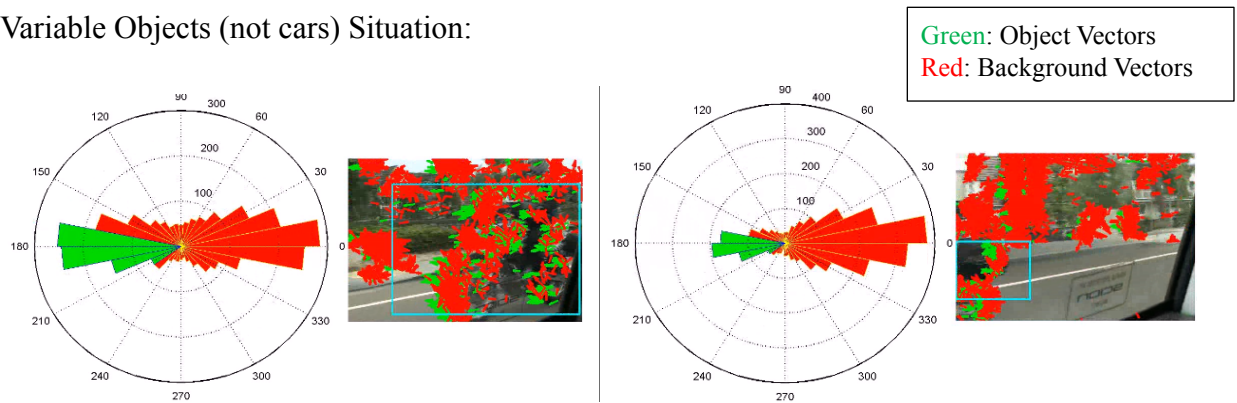

**Figure 15**: These are two images of a motorcycle in motion. Notice that it too generates the necessary green "car" vectors to create a size-fluctuating box. Thus, this shows that this optics method doesn't care about shape.

## High level methodology for detecting cars in the frontal blind spot (with no green vectors):

$$\rho = x\cos\theta + y\sin\theta$$

**Figure 16**: Hough Transform – A function that detects shapes based on a voting system.

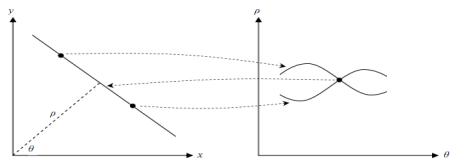

Example of Curve detection based on slope of two random points

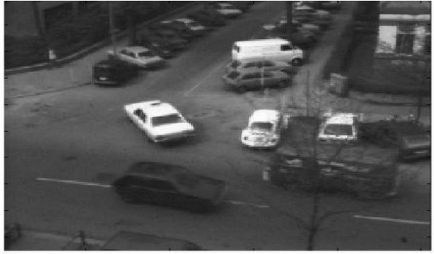 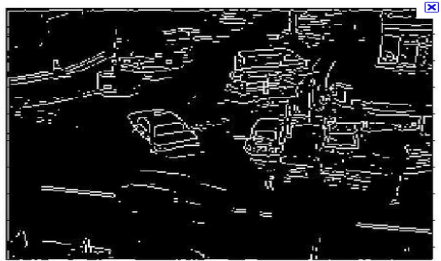

Less points to consider, which means faster computation

**Figure 17**: Edge Detection – this is an example of what edge detection does to an image frame, leaving only the important lines = less unnecessary detail

Red: Close
Orange: Still Close
Green: Medium
Blue: Far
Dark Blue: Very Far

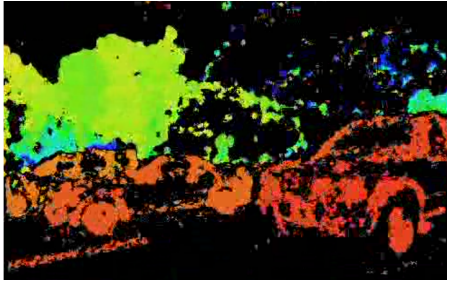 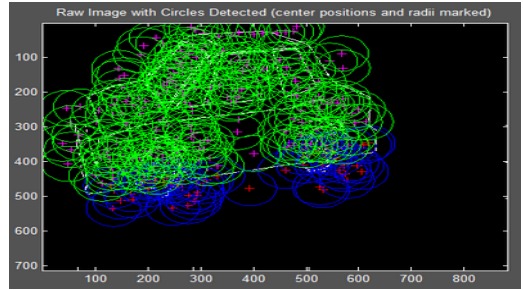

Green: non-possible wheels
Blue: Possible Wheels

**Figure 18**: Stereo Depth Filtering that shows depth based pixel color. The blue means further away, and red means closer based on a function.

**Figure 18b**: An example of a circular Hough transform on a blender generated vehicle. The blue circles are ones of interest

## Multiple Object Situations:

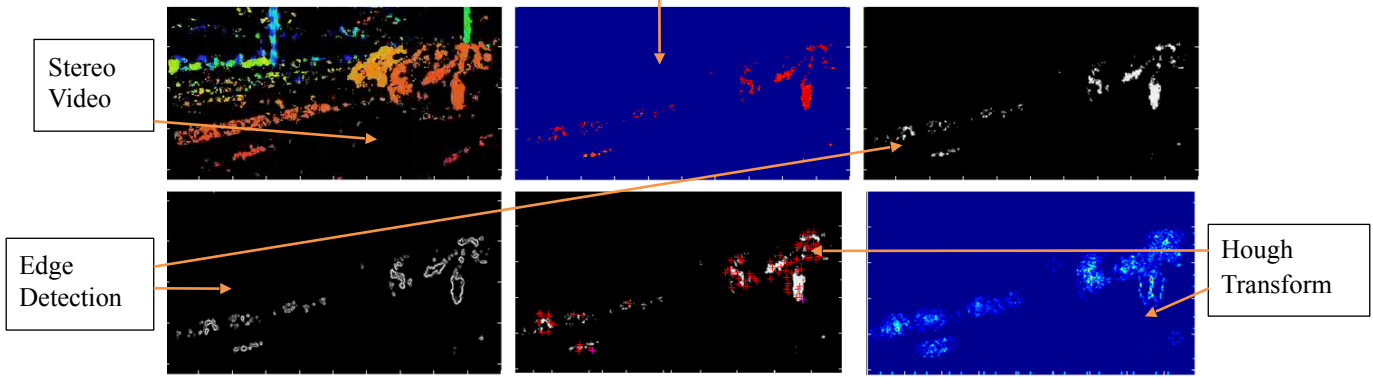

Threshold Isolation

Stereo Video

Edge Detection

Hough Transform

**Figure 19**: The stereo output is isolated to only include the vectors close by (or far away so that you can single out vehicles in a lane persay). From there, the image is put through a Gaussian gradient to produce an image of the blind spot vehicle. Edge detectors are used to nullify noise so that the car can be detected through circle (wheel) detection.

## Static Object Situation:

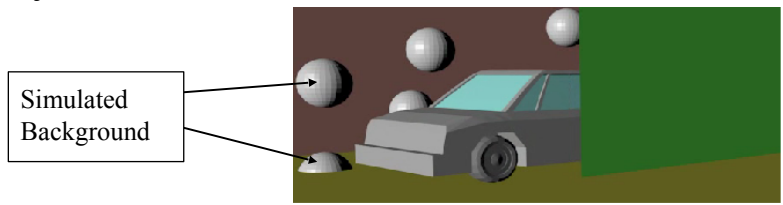

Simulated Background

**Figure 20**: Velocity of driving car = velocity of blind spot car. Must use Hough Transform and wheel detection rather than vector analysis because no green vectors.

Miscellaneous Diagrams:

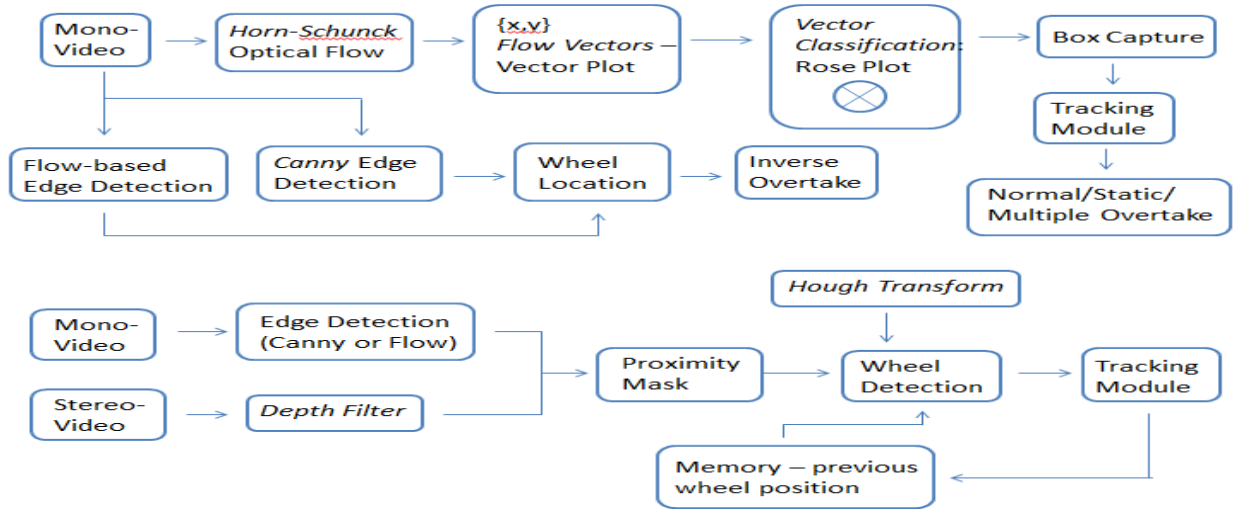

**Figure 21**: Summary of the layout of the entire project. Each individual part should be defined in the other pictures or in the written portion of methods.

## Analysis of Comparable Techniques

It is important to consider numerous ways to attack the blind spot problem. The two most common are radiofrequency and mechanical approaches.

The first attempt is through physically adding angled mirrors to cover the blind spots. Essentially, two tiny mirrors are added to the ends of the regular primary rear view mirror. The downfall is that the additional mirrors take up extra space from the frontal window that can disrupt with driving. More importantly, it can confuse the driver more as to what each mirror is intending to show.

The second, more popular approach is using radiofrequency waves to design a sort of sensor to familiarize a car with its surroundings. Although this type of approach can be adapted well for the environment, it does have crucial problems. First of all, interference is ubiquitous. Intercepting waves can distort the actual reading and alarm the driver of a "ghost car, thus provoking accidents. Two, it is extremely hard for rf tactics to distinguish the car from the background. While in optical, the directional movement pinpoints the car, it is far more difficult to read the rebounding waves and decipher which waves are pertinent to interest. Thirdly, it is important to consider that radar can only show glimpses of still framed motion (in the exact moment that the wave is shot out); it cannot account for movements of small time frames as optical fields can.

Thus, it can be argued that the addition of coding optical flow measures can be a major development in the car accident world.

## Summary and Future Work

The research proved that it is possible to have object detection based on pure visual stimulus. Although the idea of blind spot detection is relatively simple, the actual production of the source code was far more difficult than expected. The multitudes of situations that need to be covered were complicated to deal with since it required a dual system based on vector and shape detection as well as stereo implementation.

As for the validity of the results, the number of trials ran should prove that the source code produced about the same results for each video processed. Although occasionally an outlier box or an outlier shape would be detected, it did not last long enough frame-wise to create a difference or an alarm. Thus, since the source code was able to detect objects of various speeds and sizes in a number of random situations, it can be considered valid until proven faulty. However, as part of the procedure, many generic driving situations were tested; but, there may be obscure circumstances with ie) devastating weather that may cause the system to malfunction.

It would also be interesting to see how the source code would function under the abnormal conditions of angled slopes, weather extremes, etc..

As for verification in conjunction to previous work: optical flow has been used as a tracking module before in terms of traffic management and control[4]. Although the algorithms were completely different, it has been shown that it is possible to detect and track moving objects, so the results from this project should not be too surprising and should not arouse disbelief.

Besides detecting objects, another entire portion of the project that is not yet complete is the reaction system (This is will be done by the presentation date). This is based on time to collision of individual columns in an image frame. The Foci of Expansion (FOE) are the points from which the vector arrows originate from. Given an image frame at time T with the pixel (x,y,z) and the actual image location centered at the FOE with the actual pixel (X,Y,Z), congruent triangles can be compared to set $y/z = Y/Z$. But since z is a set value based on focal length of the video lens, we can set it to a constant i.e.) 1. Deriving with respect to time: $dy/dt = (dY/dt)/Z - Y((dZ/dt)/Z^2)$. Assuming $dY/dt$ is 0 (this works for whether the object is moving for stationary), substitute Y for yZ (since $y = Y/Z$) and you get $dy/dt = -y((dZ/dt)/Z)$. Dividing by a reciprocal, the result becomes $y/(dy/dt) = -Z/(dZ/dt)$, known as time to collision. Given that all the columns have their time to collisions calculated, the reaction system would be based on a heading angle. The angle starts at 0 degrees but fluctuates towards direction of the longest time to collision, thus avoiding an accident. In a blind spot situation, when cars are detected, the vehicle designated with the lowest time to collision will be detected based on columns/rows and avoided with a change in heading angle. In the sense, the goal is to have a backup machine check to see if the driver's actions in the vehicle are the smartest actions to take. If the heading angle shows a different action that may save a life, then it can override the driver.

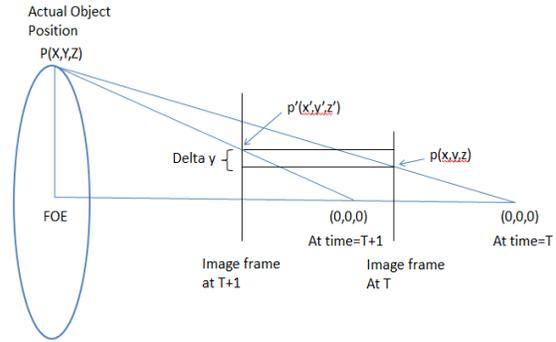

**Figure 22**: The triangles set up for the time to collision calculation. The diagram shows an image frame at two instances in time and the location of the actual object.

The most curious question that we have at this stage is under what circumstances will this source code fail and detect incorrect motion. After designing the first draft of the code, we used this same question to expand the code to cover more and more situations. However, at this time, all the particularly generic settings have been solved for. The question now lies with specific scenarios. Will a plane in the sky or even a cloud affect the vector calculations? Will small birds be detected as a car? The only way to find out is to continue testing until the motion detection system is both as accurate and fast as can be.